\documentclass[10pt, a4paper]{article}

\usepackage{lrec-coling2024} 
\usepackage{multirow}
\usepackage{hyperref}

\title{Geographically-Informed Language Identification}

\name{Jonathan Dunn$^1$ and Lane Edwards-Brown$^2$} 

\address{$^1$University of Illinois Urbana-Champaign, Department of Linguistics \\
         $^2$University of Canterbury, Department of Computer Science\\
         jedunn@illinois.edu, led43@uclive.ac.nz\\
         }

\abstract{
This paper develops an approach to language identification in which the set of languages considered by the model depends on the geographic origin of the text in question. Given that many digital corpora can be geo-referenced at the country level, this paper formulates 16 region-specific models, each of which contains the languages expected to appear in countries within that region. These regional models also each include 31 widely-spoken international languages in order to ensure coverage of these \textit{linguae francae} regardless of location. An upstream evaluation using traditional language identification testing data shows an improvement in f-score ranging from 1.7 points (Southeast Asia) to as much as 10.4 points (North Africa). A downstream evaluation on social media data shows that this improved performance has a significant impact on the language labels which are applied to large real-world corpora. The result is a highly-accurate model that covers 916 languages at a sample size of 50 characters, the performance improved by incorporating geographic information into the model.
 \\ \newline  }

\begin{document}

\maketitleabstract

\section{Geographic Priors}

Language identification (\textsc{lid}) remains a challenge for less common and low-resource languages, especially at small sample sizes (i.e., 50 characters per sample in this paper). In practical terms, the performance of \textsc{lid} models becomes a trade-off between (i) the sample size, (ii) the number of languages included, and (iii) the diversity of sources or registers per language \citep{baldwin-lui-2010-language}. Thus, large numbers of languages can be included in a single model if texts are drawn from only a single translated document like the Bible \citep{Brown2014}, but this performance does not transfer across registers \citep{lui-baldwin-2011-cross}. 

If, however, our ultimate goal is to create clean and usable corpora for less common languages, these constraints are impractical: many such corpora depend on digital registers which tend to contain short and informal texts. The basic idea in this paper is to include geographic priors about the distribution of languages in order to avoid this trade-off. More languages across more registers can be included by incorporating knowledge of the expected distribution of less common languages.

The main contribution of this paper is a \textsc{lid} model based on a fastText architecture which obtains competitive performance across 916 languages at sample sizes of 50 characters. This model is available as a Python package.\footnote{\href{https://github.com/jonathandunn/geoLid}{https://github.com/jonathandunn/geoLid}} Given the increasing number of geo-referenced corpora available \citep{Dunn2020, dunn-adams-2020-geographically}, this model thus supports higher-quality multi-lingual corpora by including more low-resource languages. 

We begin, in Section 2, by considering related work on \textsc{lid} models with a focus on specific groups of languages. The sources of data for language identification and for the geographic distribution of languages are described in Section 3 and the model architecture in Section 4. A first upstream evaluation on the \textsc{lid} task proper is undertaken in Section 5, using a held-out evaluation set as well as the smaller but more curated OpenLID data set \citep{burchell-etal-2023-open}. A second downstream evaluation is undertaken in Section 6 by applying the geographic and non-geographic \textsc{lid} models to a selection of approximately 1 million tweets each from 157 countries in order to determine the impact of this improved \textsc{lid} performance on real-world corpora. The results show that incorporating geographic information significantly improves the performance of \textsc{lid} models. More importantly, low-resource languages and under-represented populations tend to receive a larger impact from these geographically-aware models. Thus, the models improve performance in exactly those areas which are in need of such improved performance. Complete results and documentation are available in the supplementary materials for this paper.\footnote{\href{https://doi.org/10.17605/OSF.IO/RM2F3}{DOI: 10.17605/OSF.IO/RM2F3}}

\section{Related Work}

The current paper is motivated by recent work on region-specific language identification models: African languages \citep{adebara-etal-2022-afrolid}, Austronesian languages \citep{dunn-nijhof-2022-language}, Uralic languages \citep{jauhiainen-etal-2020-uralic}, Dravidian languages \citep{jauhiainen-etal-2021-comparing}, and Indo-Aryan languages \citep{jauhiainen-etal-2018-iterative}. This recent work has shown that models focused on a specific subset of languages can maintain high precision and recall while providing broad coverage within their selected domain. Although most of this work has focused on genetic groups of languages (i.e., a single language family or sub-family), there is often a close relationship between genetic and geographic groups of languages. 

At the same time, the increase in geo-referenced corpora (for example, \citealt{Dunn2020, dunn-adams-2020-geographically}) means that geographic information about the origin of samples can be considered in the corpus creation pipeline \textit{before} any \textsc{lid} model is applied. The challenge here is to incorporate geographic priors into the \textsc{lid} model itself in order to produce a larger family of region-specific language identification models, rather than the current patchwork of models for isolated groupings. While the work cited above has a narrow genetic or geographic focus, this paper expands that core idea at a global scale by training 16 region-specific models which, together, provide complete coverage of the world. 

Another reason to take a geographic approach is that the performance of \textsc{lid} models varies by population \citep{jurgens-etal-2017-incorporating}, where population can be differentiated using geographic corpora \citep{dunn-adams-2020-geographically}. This raises the challenge of whether the accuracy of \textsc{lid} models is evenly distributed across global populations, a question investigated here in the downstream evaluation in Section 6. The results suggest that \textsc{lid} models perform better for some populations than for others. The immediate implication of this finding is that digital corpora, which directly depend on \textsc{lid} models, more accurately represent some populations than others: languages which are unevenly identified cannot be correctly included in a corpus.

Another line of work has investigated the best architectures for language identification. A continued trend is that approaches based non-neural methods often out-perform neural models. Thus, there are relatively few systems that use a neural architecture \citep{kocmi-bojar-2017-lanidenn, Dunn2020, ceolin-2021-comparing} or a transformer-based architecture \citep{adebara-etal-2022-afrolid}. The most common architectures at present are a word-based back-off method (HeLI; \citealt{jauhiainen-etal-2016-heli, jauhiainen-etal-2017-evaluation, jauhiainen-etal-2022-heli}) and a character-based skip-gram approach using the fastText package \citep{joulin2016bag}. 

Given recent comparisons of these types of models \citep{dunn-nijhof-2022-language, burchell-etal-2023-open}, we employ a fastText-based architecture and focus the evaluations on geographic vs non-geographic variants. One previous issue with such architectures is a low precision for minority languages \citep{jauhiainen-etal-2022-heli}; we thus undertake a specific evaluation of minority languages in Section 5 to ensure that very common languages do not artificially inflate the performance. We also use macro rather than weighted metrics throughout the evaluation in order to avoid biasing the results towards better represented languages.

The ultimate goal of this type of \textsc{lid} model is to aid in the creation of large multi-lingual corpora, especially for less-common languages. A recent audit of such multi-lingual corpora reveals a number of mislabelled samples downstream (in the corpora) which result from mislabelled training data for \textsc{lid} models upstream \citep{kreutzer-etal-2022-quality}. We thus incorporate the findings of these audits (e.g., inconsistencies in JW 300) in order to prevent such \textsc{lid}-related issues in future corpora. The complete inventory of language codes and the language names they refer to is available in the supplementary material in order to aid future audits.\footnote{\href{https://doi.org/10.17605/OSF.IO/RM2F3}{DOI: 10.17605/OSF.IO/RM2F3}}

\section{Data Sources}

\begin{table}[h]
\centering
\begin{tabular}{|l|r|}
\hline
\textbf{Corpus} & \textbf{N. Langs} \\
\hline
Bible Translations & 614 \\
\cite{Brown2014} & ~ \\
\hline
Global Voices News & 41 \\
\cite{Tiedemann2012} & ~  \\
\hline
JW 300 & 380 \\
\cite{agic-vulic-2019-jw300} & ~ \\
\hline
Open Subtitles & 62 \\
\cite{lison-tiedemann-2016-opensubtitles2016} & ~ \\
\hline
QCRI Educational Domain & 42 \\
\cite{Tiedemann2012} & ~ \\
\hline
Tatoeba Sentences & 309 \\
\cite{Tiedemann2012} & ~ \\
\hline
Wikipedia Articles & 280 \\
TensorFlow DataSets & ~ \\
\hline
  \end{tabular}
  \caption{Primary Sources of Training Data}
  \label{tab:2}
\end{table}

The ground-truth corpora used as samples for each language are taken from several sources, detailed in Table \ref{tab:2}. Corpora are split into samples of approximately 50 characters and cleaned using the \textit{clean-text} package\footnote{\href{https://pypi.org/project/clean-text/}{https://pypi.org/project/clean-text/}} to remove \textsc{urls}, numbers, punctuation, and other non-linguistic characters. We divide the data into training, testing, and validation sets. These data sources provide a diversity of domains that is important for ensuring a robust evaluation of \textsc{lid} performance. This also provides a very large training set (over 100 million samples). Because international languages are more common and thus often contain more training samples, the number of samples for such \textit{linguae francae} is limited to 100k samples unless that language also occurs in the region in its own right. To provide a balanced evaluation, there is a limit on the number of testing samples used per source per language.

The training data described above encompasses a large number of languages but does not provide an audit of how well each sample corresponds to the given label. Previous work has shown that some datasets used for \textsc{lid} can be somewhat inaccurate \citep{kreutzer-etal-2022-quality}. Recent work has thus undertaken a similar audit of \textsc{lid} data, available as the OpenLID corpus \citep{burchell-etal-2023-open}. We use this data as an additional evaluation set for the 201 languages which it includes. This provides an additional test of whether the high performance of the geographic \textsc{lid} model extends to a smaller but cleaner test set.

\begin{table}[h]
\centering
\begin{tabular}{|ll|ll|}
\hline
\textbf{Language} & \textbf{Abbrv.} & \textbf{Language} & \textbf{Abbrv.} \\
\hline
Amharic 	&	amh	&	Korean	&	kor \\
Arabic	&	ara	&	Mandarin	&	zho \\
Bengali	&	ben	&	Marathi	&	mar \\
English	&	eng	&	Polish 	&	pol \\
Farsi	&	fas	&	Portuguese	&	por \\
French	&	fra	&	Punjabi	&	pan \\
German	&	deu	&	Russian	&	rus \\
Gujarati 	&	guj	&	Spanish	&	spa \\
Hausa	&	hau	&	Swahili	&	swa \\
Hindi	&	hin	&	Tagalog	&	tgl \\
Indonesian	&	ind	&	Tamil	&	tam \\
Italian 	&	ita	&	Telugu	&	tel \\
Japanese	&	jpn	&	Thai 	&	tha \\
Javanese	&	jav	&	Turkish	&	tur \\
Kannada 	&	kan	&	Urdu	&	urd \\
~	&	~	&	Vietnamese 	&	vie \\
\hline
  \end{tabular}
  \caption{International languages which are included in each regional model.}
  \label{tab:international}
\end{table}

\begin{table*}[t]
\centering
\begin{tabular}{|lc|cc|cc|cc|r|}
\hline
\multirow{2}{*}{\textbf{Region}} & \textbf{N.} & \multicolumn{2}{c|}{\textbf{Precision}} & \multicolumn{2}{c|}{\textbf{Recall}} & \multicolumn{2}{c|}{\textbf{F-Score}} & \multicolumn{1}{c|}{\textbf{Test}}  \\
~ & \textbf{Langs} & \textit{Geo} & \textit{Baseline} & \textit{Geo} & \textit{Baseline} & \textit{Geo} & \textit{Baseline} & \textbf{Samples} \\
\hline
Africa, North & 44 & \textbf{0.993} & 0.838 & \textbf{0.988} & 0.974 & \textbf{0.990} & 0.886 & 621k \\
Africa, Southern & 58 & \textbf{0.983} & 0.866 & \textbf{0.983} & 0.972 & \textbf{0.982} & 0.903 & 1,053k \\
Africa, Sub & 166 & \textbf{0.981} & 0.936 & \textbf{0.981} & 0.971 & \textbf{0.980} & 0.947 & 1,931k \\
\hline
America, Central & 188 & \textbf{0.991} & 0.950 & \textbf{0.991} & 0.983 & \textbf{0.991} & 0.961 & 2,965k \\
America, North & 68 & \textbf{0.993} & 0.869 & 0.991 & 0.963 & \textbf{0.993} & 0.902 & 1,017k \\
America, South & 129 & \textbf{0.994} & 0.943 & \textbf{0.996} & 0.990 & \textbf{0.995} & 0.960 & 4,612k \\
America, Brazil & 88 & \textbf{0.996} & 0.919 & \textbf{0.996} & 0.990 & \textbf{0.996} & 0.945 & 818k \\
\hline
Asia, Central & 54 & \textbf{0.988} & 0.867 & \textbf{0.988} & 0.978 & \textbf{0.988} & 0.906 & 777k \\
Asia, East & 46 & \textbf{0.991} & 0.848 & \textbf{0.990*} & 0.976 & \textbf{0.990} & 0.892 & 1,131k \\
Asia, South & 60 & \textbf{0.988} & 0.882 & \textbf{0.985*} & 0.978 & \textbf{0.986} & 0.914 & 979k \\
Asia, Southeast & 325 & \textbf{0.990} & 0.968 & \textbf{0.990} & 0.983 & \textbf{0.990} & 0.972 & 3,992k \\
\hline
Europe, East & 65 & \textbf{0.982} & 0.876 & \textbf{0.976} & 0.968 & \textbf{0.978} & 0.908 & 3,132k \\
Europe, West & 108 & \textbf{0.970} & 0.903 & \textbf{0.964} & 0.958 & \textbf{0.967} & 0.921 & 5,473k \\
Europe, Russia & 65 & \textbf{0.985} & 0.884 & \textbf{0.981} & 0.972 & \textbf{0.984} & 0.914 & 1,098k \\
\hline
Middle East & 53 & \textbf{0.989} & 0.866 & \textbf{0.986} & 0.977 & \textbf{0.988} & 0.904 & 801k \\
Oceania & 49 & \textbf{0.988} & 0.854 & \textbf{0.980} & 0.965 & \textbf{0.984} & 0.890 & 745k \\
\hline
  \end{tabular}
  \caption{Performance of the geographically-aware language identification model against a baseline model trained with the same data, organized by region. Each sample is approximately 50 characters in length. All \textbf{bold} values are significantly better than the baseline using a t-test paired by language; values marked with \textbf{*} are significant at the $p<0.01$ level and all others at the $p<0.001$ level. All values are reported using the Macro-Average to avoid privileging well-represented languages.}
  \label{tab:1}
\end{table*}

\textbf{Geographic Distribution of Languages}. This paper takes a country-level approach to geographic \textsc{lid} in which the country from which a sample originates determines the set of languages it is expected to represent. Countries are then aggregated into 16 larger regions, as listed in Table \ref{tab:1}, such as North America or the Middle East. Because there are many possible connections within regions, any language which appears in one country is expected to sometimes appear in other countries within the same region. This assumption prevents the model from maintaining an overly narrow definition of the distribution of languages.

We distinguish between \textsc{local} and \textsc{international} languages. The latter are widely-spoken \textit{linguae francae} which are expected to appear everywhere. Here, we include 31 international languages, as listed in Table \ref{tab:international}; these are selected either because of their number of speakers or because they appear independently in several regions. These international languages, then, are included in each region-specific model regardless of whether they specifically appear in that region. The idea behind this choice is that human populations move around because of immigration and tourism; such populations are expected to rely on these \textit{linguae francae} in their new locations.

This contrasts with \textsc{local} languages which are only present in a region-specific model if they are expected to appear in that region. Various sources were considered as ground-truth information for the geographic distribution of languages, starting with estimates like the CIA World Factbook. However, the most authoritative data comes from Glottolog\footnote{\href{https://github.com/glottolog/glottolog}{https://github.com/glottolog/glottolog}}, which provides the primary source of geographic information \citep{Glotto}. The database which maps between language codes, countries, and region-specific models is available in the supplementary material.\footnote{\href{https://doi.org/10.17605/OSF.IO/RM2F3}{DOI: 10.17605/OSF.IO/RM2F3}} The total number of languages within a region ranges from 44 (North Africa) to 325 (Southeast Asia). 

Rather than weight each language by its proportion of speakers in a region we consider all languages within a regional model to be equally likely. This is a practical choice given the unreliability of such estimates for many countries. For example, one potential weighting scheme would be to privilege languages used by the majority of the population in a specific country using estimates of the percentage of the population which use each language. In practice, these estimates proved to be inconsistent, especially for under-represented populations with a greater need for more accurate \textsc{lid}. Another potential weighting scheme would be to restrict the languages available per country rather than per region. Again, however, this requires more precise information about what languages are actually used in each country than is currently available. In a region-based approach we are able to share information across neighbouring countries and thus improve performance for all countries in that region.

On the other hand, this region-based approach places greater importance on the division of countries into regional groups. Here we have followed the division used in the earthLings.io language mapping project for convenience.\footnote{\href{https://www.earthlings.io}{https://www.earthLings.io}} The only challenge arises when a country sits at the intersection between two regions, such as Iran with the Middle East and Central Asia. These challenges have been mitigated here by adopting a broad set of international languages; one criteria is the number of speakers per language but a second criteria is the number of regions which a language occurs in. When languages found in the border between regions are present in all relevant models, the danger of a region-based model misplacing those languages is reduced. From an empirical perspective, the improved performance reported below shows that even region-based geographic priors are sufficient for more accurate \textsc{lid}.

\section{Models}

GeoLID is comprised of 16 distinct models, one for each geographic region as listed in Table \ref{tab:1}. In addition, we train a single baseline model using the same architecture and training data but without any geographic information. This allows us to compare the impact which is derived specifically from geographic priors rather than other areas of improvement. In the Python package\footnote{\href{https://github.com/jonathandunn/geoLid}{https://github.com/jonathandunn/geoLid}}, this baseline model is used for cases in which no geographic information is available.

\begin{table*}[t]
\centering
\begin{tabular}{|lc|cc|cc|cc|r|}
\hline
\multirow{2}{*}{\textbf{Region}} & \textbf{N.} & \multicolumn{2}{c|}{\textbf{Precision}} & \multicolumn{2}{c|}{\textbf{Recall}} & \multicolumn{2}{c|}{\textbf{F-Score}} & \multicolumn{1}{c|}{\textbf{Test}}  \\
~ & \textbf{Langs} & \textit{Geo} & \textit{Baseline} & \textit{Geo} & \textit{Baseline} & \textit{Geo} & \textit{Baseline} & \textbf{Samples} \\
\hline
Africa, North & 13 & \textbf{0.992} & 0.976 & \textbf{0.982} & 0.965 & \textbf{0.986} & 0.972 & 75k \\
Africa, Southern & 27 & \textbf{0.980} & 0.966 & \textbf{0.974} & 0.964 & \textbf{0.976} & 0.964 & 442k \\
Africa, Sub & 135 & \textbf{0.980} & 0.972 & \textbf{0.979} & 0.969 & \textbf{0.979} & 0.970 & 1,313k \\
\hline
America, Central & 157 & \textbf{0.991} & 0.983 & \textbf{0.991} & 0.984 & \textbf{0.991} & 0.983 & 2,339k \\
American, North & 37 & \textbf{0.996} & 0.943 & 0.991 & 0.951 & \textbf{0.994} & 0.945 & 363k \\
America, South & 98 & \textbf{0.997} & 0.994 & \textbf{0.997} & 0.994 & \textbf{0.997} & 0.994 & 3,747k \\
America, Brazil & 57 & \textbf{0.999} & 0.995 & \textbf{0.999} & 0.996 & \textbf{0.999} & 0.996 & 287k \\
\hline
Asia, Central & 23 & 0.986 & 0.985 & \textbf{0.985} & 0.979 & 0.984 & 0.983 & 237k \\
Asia, East & 15 & \textbf{0.996} & 0.988 & 0.989 & 0.972 & 0.993 & 0.980 & 410k \\
Asia, South & 29 & 0.990 & 0.991 & 0.980 & 0.979 & 0.983 & 0.983 & 340k \\
Asia, Southeast & 294 & \textbf{0.992} & 0.988 & \textbf{0.990} & 0.984 & \textbf{0.991} & 0.985 & 3,291k \\
\hline
Europe, East & 34 & \textbf{0.985} & 0.964 & \textbf{0.968} & 0.959 & \textbf{0.975} & 0.961 & 2,363k \\
Europe, West & 77 & \textbf{0.971} & 0.952 & \textbf{0.957} & 0.950 & \textbf{0.963} & 0.950 & 4,565k \\
Europe, Russia & 34 & \textbf{0.983} & 0.979 & \textbf{0.974} & 0.967 & \textbf{0.979} & 0.972 & 525k \\
\hline
Middle East & 22 & 0.992 & 0.986 & \textbf{0.980} & 0.975 & \textbf{0.986} & 0.981 & 255k \\
Oceania & 18 & \textbf{0.994} & 0.983 & \textbf{0.962} & 0.943 & \textbf{0.978} & 0.959 & 168k \\
\hline
  \end{tabular}
  \caption{Performance of the geographically-aware language identification model against a baseline model trained with the same data, organized by region. \textbf{This table shows results for only local languages in each region to avoid inflating the performance with high-resource languages}. Each sample is approximately 50 characters in length. All \textbf{bold} values are significantly better than the baseline using a t-test paired by language at the $p<0.05$ level. All values are reported using the Macro-Average.}
  \label{tab:local}
\end{table*}

Given the results of previous evaluations, all models are based on the fastText architecture \citep{dunn-nijhof-2022-language}. Parameters are determined using experiments on a development set. In particular, the skip-gram architecture is used with a negative sample rate of 100 and an increase in n-gram buckets to 4 million. Initial experiments showed that languages with inconsistent word segmentation practices (such as Chinese or Japanese) performed inconsistently. This was corrected by adding whitespace between all characters regardless of language and increasing the word n-gram limit to 6; effectively, this variant ignores word boundaries regardless of language and thus better captures those languages in which word boundaries vary by corpus. Experiments on model compression \citep{joulin2016fasttextzip} show that such models have significantly reduced performance; these comparisons are available in the supplementary material but not described in the main text.

\section{Upstream Evaluation}

\begin{table}[h]
\centering
\begin{tabular}{|l|c|r|r|}
\hline
\textbf{Region} & \textbf{N.} & \textbf{Geo} & \textbf{Baseline} \\
\hline
Africa, North & 35 & \textbf{0.990}	&	0.969 \\
Africa, Southern & 44 & \textbf{0.988}	&	0.968 \\
Africa, Sub & 65 & \textbf{0.983}	&	0.968 \\
\hline
America, Central & 36 & \textbf{0.987}	&	0.970 \\
America, North & 31 & \textbf{0.986}	&	0.966 \\
America, South & 35 & \textbf{0.990}	&	0.969 \\
America, Brazil & 33 & \textbf{0.993}	&	0.967 \\
\hline
Asia, Central & 43 & \textbf{0.991}	&	0.974 \\
Asia, East & 38 & 0.964	&	0.945 \\
Asia, South & 42 & 0.947	&	0.925 \\
Asia, Southeast & 48 & 0.936	&	0.926 \\
\hline
Europe, East & 53 & \textbf{0.992}	&	0.978 \\
Europe, West & 67 & \textbf{0.991} &	0.982 \\
Europe, Russia & 42 & \textbf{0.992} &	0.974 \\
\hline
Middle East & 41 & \textbf{0.994}	&	0.973 \\
Oceania & 35 & 0.976 &	0.960 \\
\hline
  \end{tabular}
  \caption{Evaluation using OpenLID Data. The Macro F-Score is reported to ensure equal contribution across languages. Values in bold are significantly higher at the $p<0.05$ level using a one-tailed t-test paired by language. There are fewer observed languages per region because the OpenLID data includes fewer languages overall.}
  \label{tab:3}
\end{table}

The first evaluation uses traditionally-labelled \textsc{lid} corpora to evaluate performance of the regional models against the non-geographic baseline. The number of test samples per language is limited to 15k to avoid giving preference to majority languages, which would support a very large number of samples; no more than 2k samples come from a single source. In addition, we report the macro-average for each region rather than the weighted-average, which further avoids privileging well-represented languages. 

The first part of this evaluation, on a held-out evaluation set, is shown in Table \ref{tab:1}. The performance is broken down by region, with each region evaluated on those languages which are present in that area; the performance of the baseline model (which includes all 916 languages) is here limited to that same set of region-specific languages. The number of languages per region is also shown in the table. Performance is not dictated by the number of languages; for example, the f-score for Southeast Asia (with 325 languages) is 0.990 while the f-score for Western Europe (with 108 languages) is somewhat lower at 0.967. This indicates that the mix of languages in Western Europe is more difficult even with fewer of them overall.

The significance of the differences between the geographic model and the baseline model are calculated using a paired two-tailed t-test; almost all values are significant at the $p<0.001$ level, with two significant at only the $p<0.01$ level and one (recall in North America) with no significant difference. Performance remains high even with a large number of test samples; the lowest f-score is 0.967 in Western Europe; only Western and Eastern Europe fall below an f-score of 0.980. In the aggregate, then, this shows the advantage of tailored \textsc{lid} models for each region.

Given previous audits of multi-lingual resources, our next question is whether this performance is restricted to this particular evaluation set (a held-out portion of the same corpora used for training). Thus, we use the independently audited OpenLID data for an additional evaluation, using all samples above 50 characters. This data set contains only 201 languages, so this serves as a quality-control check for the more common languages. As before, we compare the geographic model for each region against the baseline model. This is shown in Table \ref{tab:3}, with differences significant at the $p<0.05$ marked in bold; as before, this shows the macro-average f-score. First, we see that the results are generally comparable with the previous evaluation; in fact, most regions obtain f-scores above 0.990. The regions which deviate from this generalization are also not significantly better in the geographic model than the baseline mode: East, South, and Southeast Asia and Oceania. It should be noted that these results encompass only 201 languages, which means that this performance is more focused on more widely available languages.

The source of the lower performance in these four regions is a small number of languages whose representation in the training set is not similar to the OpenLID set; these are potentially mislabelled languages. This comes down to five individual languages: Banjar (bjn), Maithili  (mai), Burmese (mya), Sanskrit (san), and Shan (shn). With the exception of these five languages, then, the evaluation on the OpenLID data validates the previous evaluation on a sub-set of the languages.\footnote{For the final released models, these five languages are retrained by incorporating OpenLID as training data to ensure valid representations.}

\begin{table}[h]
\centering
\begin{tabular}{|l|c|c|c|}
\hline
\textbf{Model} & \textbf{Lang.} & \textbf{Geo} & \textbf{Non-Geo} \\
\hline
Africa, Southern & sot & 0.88 & 0.86 \\
Africa, Sub & bam & 0.69 & 0.53 \\
Africa, Sub & fuh & 0.69 & 0.58 \\
Africa, Sub & ffm & 0.70 & 0.59 \\
Africa, Sub & plt & 0.86 & 0.85 \\
Africa, Sub & tum & 0.86 & 0.83 \\
Africa, Sub & eng & 0.87 & 0.34 \\
Africa, Sub & run & 0.89 & 0.89 \\
America, Central & kek & 0.87 & 0.85 \\
Asia, South & dty & 0.79 & 0.80 \\
Asia, Southeast & pam & 0.81 & 0.76 \\
Asia, Southeast & cbk & 0.85 & 0.81 \\
Asia, Southeast & spa & 0.85 & 0.41 \\
Asia, Southeast & tet & 0.85 & 0.78 \\
Asia, Southeast & bjn & 0.86 & 0.82 \\
Asia, Southeast & gor & 0.87 & 0.84 \\
Europe, East & eng & 0.87 & 0.34 \\
Europe, East & rmy & 0.87 & 0.75 \\
Europe, Russia & mdf & 0.89 & 0.87 \\
Europe, West & eng & 0.86 & 0.34 \\
Europe, West & ile & 0.87 & 0.88 \\
Oceania & cha & 0.86 & 0.78 \\
Oceania & bjn & 0.88 & 0.82 \\
\hline
  \end{tabular}
  \caption{Complete list of all languages with an f-score below 0.90 in the geographic models.}
  \label{tab:low}
\end{table}

As noted above, one potential issue with evaluating \textsc{lid} models is that the high performance of very common international languages (like English or French) could inflate the overall performance of the model. In the case of corpus creation, however, it is often the less-common languages which we are more interested in. Our first approach to dealing with this issue is (i) to limit the number of test samples to 15k per language and (ii) report the macro- rather than weighted-average. Here we undertake a further analysis of the performance of the models by region, as shown in Table \ref{tab:local}, with the 31 international languages excluded from the results. Thus, this table shows the performance of the models by region while focusing on the less-common languages from each area.

The main conclusion from Table \ref{tab:local} is that the high performance of these models is not dependent on a few high-performing and very common languages. The f-scores are in some cases lower when we focus on less-common languages, but the difference is minor. For example, the lowest performing region Western Europe declines from 0.967 to 0.963. In most cases the difference between the geographic model and the baseline model remains significant; where this is not the case, it is partly because the number of languages being compared is now lower overall (i.e., only 15 in East Asia). This table indicates, then, that the high performance of these regional \textsc{lid} models is not an artifact of having many test samples from high-resource languages.

It is still possible that these macro-averages disguise low-performing languages in the models. We thus conduct a check for all languages with an f-score below 0.85. The main regional model with cases of low performance is Sub-Saharan Africa.  There are three such languages here with f-scores below 0.80: Bambara (bam), Maasina Fulfulde (ffm), Western Niger Fulah (fuh). The latter two are quite closely related. One language in South Asia, Dotyali
 (dty), also falls below an f-score of 0.80. This analysis shows, then, that these models retain high performance for 912 of 916 languages.

To further illustrate this point, the complete list of all languages with an f-score below 0.90 in any geographic model is shown in Table \ref{tab:low}. The geographic f-score is compared with the non-geographic f-score. Even in cases where performance is relatively low, there is still a stronger performance in the models with geographic information. Taken as a whole, this line of evaluation shows how geographic information can improve performance specifically for under-represented languages. The final models have consistently strong performance across almost all languages.

\section{Downstream Evaluation}

The evaluation has so far focused on traditional \textsc{lid} data that is annotated with language labels and drawn from more formal sources like the Bible and Wikipedia. The region-specific models perform significantly better on this data in such an upstream evaluation. The remaining question is whether a few points of f-score in an upstream evaluation make a significant difference downstream on the kind of informal digital texts that are used to build large corpora. Here we evaluate this using geo-referenced tweets. The evaluation corpus contains 189 million tweets drawn equally from 157 countries; each tweet contains at least 50 characters after cleaning. Both the geographic model and the baseline model are used to annotate this corpus. The question is whether the improvements in the geographic model upstream have an impact on real-world datasets.

\begin{figure*}[t]
\centering
\includegraphics[width = 450pt]{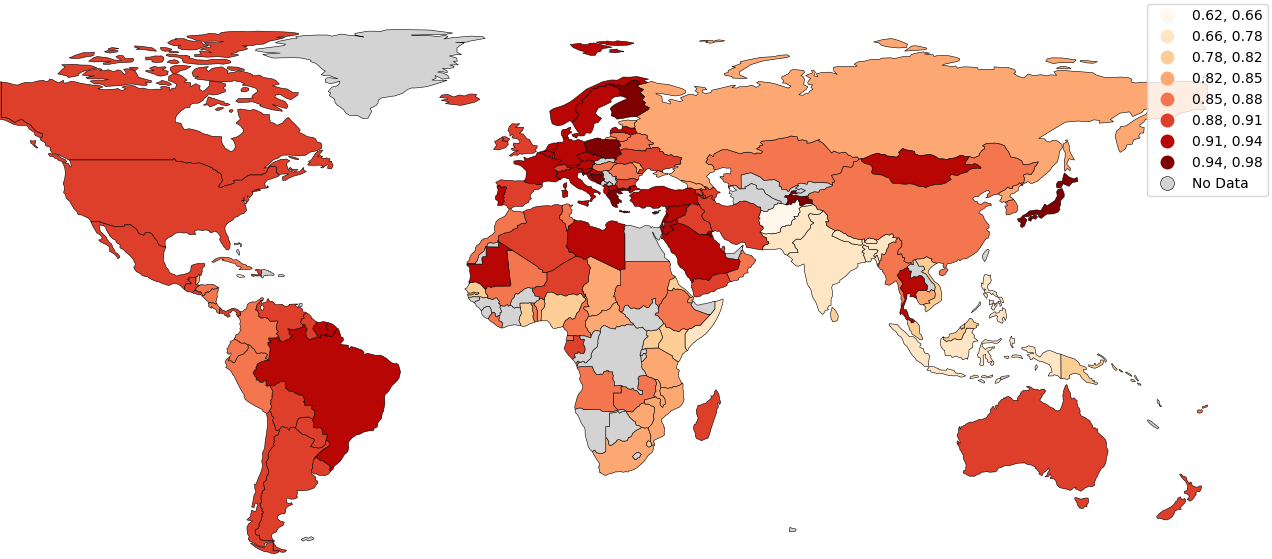}
\caption{Map showing agreement between language identification models by country. A value of 0.80 means that 80\% of samples receive the same language label from each model. Agreement is calculated using approximately 1 million random tweets per country, where each tweets has at least 50 characters.}
\label{fig:map}
\end{figure*}

\begin{table}[h]
\centering
\begin{tabular}{|l|c|r|r|}
\hline
\textbf{Region} & \textbf{N.} & \textbf{Agree} & \textbf{Samples} \\
\hline
Africa, North & 9 & 84.93\% & 10.21 mil \\
Africa, Southern & 3 & 83.08\% & 3.77 mil \\
Africa, Sub & 26 & 84.94\% & 26.71mil \\
\hline
America, Central & 14 & 87.28\% & 16.71 mil \\
America, North & 2 & 89.62\% & 2.97 mil \\
America, South & 11 & 88.85\% & 14.40 mil \\
America, Brazil & 1 & 92.41\% & 1.53 mil \\
\hline
Asia, Central & 7 & 85.52\% & 9.28 mil \\
Asia, East & 5 & 89.31\% & 5.96 mil \\
Asia, South & 7 & 74.20\% & 8.42 mil \\
Asia, Southeast & 14 & 83.86\% & 13.39 mil \\
\hline
Europe, East & 15 & 89.88\% & 21.20 mil \\
Europe, West & 23 & 91.03\% & 32.19 mil \\
Europe, Russia, & 1 & 83.03\% & 1.50 mil \\
\hline
Middle East & 12 & 91.80\% & 16.33 mil \\
Oceania & 7 & 85.56\% & 4.75 mil \\
\hline
\textbf{Total} & \textbf{157} & \textbf{87\%} & \textbf{189 mil} \\
\hline
  \end{tabular}
  \caption{Agreement by region of the language labels predicted for 189 million tweets from 157 countries. \textit{Agreement} here is the percentage of samples for which the geographic model and the comparable non-geographic baseline model predict the same language label. All samples contain at least 50 characters. \textit{N} here refers to the number of individual countries per region.}
  \label{tab:agreement}
\end{table}

The aggregated agreement rates by region are shown in Table \ref{tab:agreement}. While the average agreement is 87\%, this ranges from a low of 74\% in South Asia to a high of 92\% in Brazil. This is our first indication that the impact of a geographic approach to \textsc{lid} has varying impacts in different locations. The more important point, though, is that even a few points of difference in f-score in the upstream evaluation have a significant impact on the make-up of a downstream corpus. Why? First, of course, is the large number of samples in this geographic data set. Small differences in precision and recall, when multiplied on this scale, influence a large number of samples. Second, this kind of informal digital text is quite different from the formal registers available in traditional \textsc{lid} training sets, which means that accurate identification is more difficult. Given the more difficult context, the increased performance from incorporating geographic information becomes even more consequential.

\begin{table*}[t]
\centering
\begin{tabular}{|lcr|lcr|lcr|}
\hline
\multicolumn{3}{|c|}{\textbf{High Agreement}} & \multicolumn{3}{|c|}{\textbf{Medium Agreement}} & \multicolumn{3}{|c|}{\textbf{Low Agreement}}  \\
\multicolumn{1}{|c}{\textit{Language}} & \textit{\%} & \textit{N.} & \multicolumn{1}{|c}{\textit{Language}} & \textit{\%} & \textit{N.} & \multicolumn{1}{|c}{\textit{Language}} & \textit{\%} & \textit{N.} \\
\hline
Sinhala  (sin) & 99\% & 204k & Polish (pol) & 79\% & 2065k & Somali (som) & 56\% & 483k \\ 
Hebrew (heb) & 99\% & 752k & Farsi (fas) & 79\% & 2110k & Scots (sco) & 49\% & 201k \\ 
Thai (tha) & 98\% & 1508k & Ukrain. (ukr) & 77\% & 1202k & Banjar. (bjn) & 47\% & 107k \\ 
Maced.  (mkd) & 98\% & 805k & Uzbek (uzb) & 76\% & 232k & Hindi (hin) & 45\% & 1262k \\ 
Japanese (jpn) & 98\% & 4019k & Shona (sna) & 76\% & 313k & Sundan. (sun) & 44\% & 168k \\ 
Swedish (swe) & 98\% & 869k & Indon. (ind) & 74\% & 4950k & Wolof (wol) & 41\% & 201k \\ 
Arabic (ara) & 98\% & 14585k & Chewa (nya) & 72\% & 133k & Bavarian (bar) & 39\% & 265k \\ 
Latvian (lav) & 97\% & 974k & Swahili (swa) & 71\% & 1688k & Hausa (hau) & 38\% & 297k \\ 
Maldivian  (div) & 97\% & 159k & Luxemb. (ltz) & 68\% & 106k & G. Konk. (gom) & 26\% & 195k \\ 
Greek (ell) & 97\% & 1422k & Kurdish (kur) & 68\% & 152k & Slovak (slk) & 24\% & 172k \\ 
Korean (kor) & 96\% & 1225k & Estonian (est) & 68\% & 244k & Venetian (vec) & 21\% & 114k \\ 
Amharic (amh) & 95\% & 149k & Albanian (sqi) & 67\% & 738k & Javanese (jav) & 20\% & 1424k \\ 
Armenian  (hye) & 95\% & 294k & Vietn. (vie) & 67\% & 443k & Malagasy (mlg) & 15\% & 264k \\ 
Finnish (fin) & 95\% & 1234k & Tagalog (tgl) & 61\% & 2279k & Quech. (que) & 14\% & 192k \\ 
Mong. (mon) & 94\% & 362k & Zulu (zul) & 58\% & 112k & Afrikaans (afr) & 11\% & 522k \\ 
\hline
  \end{tabular}
  \caption{Average agreement across countries for select languages. \textit{Agreement} is the percent of cases in which the geographic model and the baseline model assign the same label. Only languages with at least 1 million samples are considered.}
  \label{tab:agreement_avg}
\end{table*}

For this downstream evaluation we do not know the ground-truth labels for each sample. What we do know, however, is that the geographic models have higher precision and recall on the upstream evaluation, meaning that they are better able to correctly identify the language of samples. On this region-specific data the baseline model must consider all languages even though, given geographic information, we know that less common local languages in one place are unlikely to occur in another place. On a large scale, this prior information makes a large impact on the make-up of an automatically-created corpus; on average, over 13\% of the samples would be labelled differently depending on the model used. This is a meaningful difference when we consider that low-resource languages constitute much less than 13\% of digital language data.

Another perspective to the agreement between the geographic model and the baseline model is to organize the results by language. This is shown in Table \ref{tab:agreement_avg}, where the language label is derived from the more accurate geographic model. \textit{Agreement} is thus the percentage of cases in which the baseline also predicts the same label. It is possible, of course, that both models are wrong. In cases of disagreement, however, we are sure that at least one model is wrong and most likely that is the baseline model. The table shows a selection of languages with high agreement (left), medium agreement (center), and low agreement (right).

On the left in Table \ref{tab:agreement_avg} are languages for which the geographic approach to \textsc{lid} would have little impact: both models largely agree. Thus, a corpus produced using these models in the pipeline would be much the same in either case. These tend to be more common languages, like Thai or Japanese or Arabic. However, some less common languages like Armenian and Sinhala also have this high agreement. For these cases, then, there is less impact downstream from the improvements provided by the geographic models.

In the center in Table \ref{tab:agreement_avg} are languages with significantly lower agreement. Surprisingly, many of these languages are also quite common, like Farsi and Indonesian. Others, like Zulu or Uzbek, are somewhat less common. In either case, any corpora depending on a non-geographic model (the baseline) would be significantly different from the more accurate geographic variant. Thus, in these cases the impact is quite meaningful.

Finally, on the right in Table \ref{tab:agreement_avg} are languages with low agreement, down to as low as 11\% and 15\%. These are mostly less common languages like Hausa or Wolof, but with some very common cases like Hindi (at only 45\% agreement). For these languages, there would be very little overlap in the downstream corpora created by the two \textsc{lid} models. Thus, in these cases there is a highly meaningful difference created by incorporating geographic priors into the \textsc{lid} component.

The Herfindahl–Hirschman Index (HHI) for these low-agreement languages is shown in Table \ref{tab:hhi}. This is a measure of monopoly adapted from economics which is calculated by taking the sum of the square of shares of error labels. Thus, if Javanese has a low agreement because the non-geographic model always predicts this to be Indonesian, the HHI would be high (showing a monopoly of errors). A low value, however, means that the baseline model's predictions are spread across many languages rather than just Indonesian. This table shows that the low agreement is not driven by the baseline model over-predicting a single language.

\begin{table}[h]
\centering
\begin{tabular}{|lc|}
\hline
\textbf{Language} & \textbf{Hirschman Index} \\
\hline
Somali (som) & 0.10 \\
Scots (sco) & 0.29 \\
Banjarese (bjn) & 0.14 \\
Hindi (hin) & 0.14 \\
Sundanese (sun) & 0.17 \\
Wolof (wol) & 0.16 \\
Bavarian (bar) & 0.19 \\
Hausa (hau) & 0.09 \\
G. Konk. (gom) & 0.09 \\
Slovak (slk) & 0.10 \\
Venetian (vec) & 0.13 \\
Javanese (jav) & 0.07 \\
Malagasy (mlg) & 0.06 \\
Quechuan (que) & 0.09 \\
Afrikaans (afr) & 0.06 \\
\hline
  \end{tabular}
  \caption{Herfindahl–Hirschman Index for languages with low agreement. A high value indicates that most cases of disagreement are given a single label while a low value means that cases of disagreement are widely distributed across labels.}
  \label{tab:hhi}
\end{table}

The purpose of the downstream analysis in this section has been to determine whether the significant but still small differences in upstream prediction performance have an impact on a large, real-world corpus. This corpus of tweets is much less formal than the traditional \textsc{lid} training data and represents a much broader range of populations, since it has been collected from 157 different countries. The results show quite clearly that there is a significant downstream impact, with some places (e.g., Table \ref{tab:agreement} and Figure \ref{fig:map}) and some languages (e.g., Table \ref{tab:agreement_avg}) being more influenced than others. This is important because it means that the improvements made upstream in the modelling task have an impact downstream.

Consider the map of agreement rates by country in Figure \ref{fig:map}, averaged across languages. Places that are a darker red, like North America or Western Europe, have generally higher agreement. This means that there is less of an impact from geographic priors on language identification and thus on downstream corpus creation. These places and populations are better represented in the digital world. Places which are lighter red, however, like India or Malaysia, have much lower agreement. This means, conversely, that there is a greater impact here in incorporating geographic priors into language identification. These populations are less well represented in the digital world and many of the languages here are low-resource languages. It is precisely these under-represented populations and low-resource languages which benefit from a geographic approach to \textsc{lid}.

\section{Discussion and Conclusions}

The main contribution of this paper has been to explore whether the incorporation of known geographic information about the distribution of languages into a language identification model improves performance. The idea is that less-common or low-resource languages are usually local to particular parts of the world. The approach taken in this paper has been to distinguish between international languages which are widely spoken and may occur anywhere (31 in total) and local languages which are expected to occur in specific regions (885 in total). By formulating 16 region-specific models, this approach is able to include more languages while maintaining a higher prediction accuracy. The use of regional geographic information rather than more precise country-level geographic information compensates for the fact that census-level information about language use is inconsistent across many parts of the world.

The first set of evaluations, upstream, evaluated the models using traditional \textsc{lid} data. This evaluation showed that geographic models have significantly better performance than a baseline model of the same architecture but without geographic information. The evaluation also incorporated independent and curated data from the OpenLID dataset to validate the performance of these models. Further, an investigation of less-common languages shows that only four out of 885 have a low precision and recall which would call into question corpora created with the output of the models.

The second set of evaluations, downstream, used a corpus of 189 million tweets from 157 countries to determine whether the upstream differences in prediction accuracy have an impact on the predicted labels on a real-world corpus. These results show an average agreement of 87\%, meaning that 13\% of the data is handled differently by the more-accurate geographic \textsc{lid} models. Importantly, though, the analysis by country and by language shows that disagreement is not evenly distributed. This means that the impact of geographic \textsc{lid} models is strongest in precisely those contexts which need higher quality corpora. Non-geographic \textsc{lid} models perform worst in low-resource contexts for which accurate corpus creation is most needed.

Not all sources of data can be made to benefit from these improved models. However, many large digital corpora, from the web and from social media, can be geo-referenced \citep{Dunn2020, dunn-adams-2020-geographically}. The earthLings.io language mapping project, for example, has shown how much geo-referenced corpus data is available from digital sources.\footnote{\href{https://www.earthlings.io}{https://www.earthLings.io}} These large corpora, with the addition of improved geographically-informed language identification, can then be used to create corpora which better represent the languages and language varieties used by under-represented portions of the world's population. The contribution of this paper is two-fold: first, a systematic evaluation of the geographic distribution of the performance of \textsc{lid} models and, second, the release of the resulting GeoLID models\footnote{\href{https://github.com/jonathandunn/geoLid}{https://github.com/jonathandunn/geoLid}} which increase both (i) the number of languages considered as well as (ii) the performance on individual languages.

\section{Ethics Statement}

This paper relies on existing multi-lingual data sets for training and evaluating language identification models. While every effort has been made to correct for known inconsistencies and to use external validation data, the authors are unable to personally evaluate the quality of the models across all 916 languages included.

\section{Bibliographical References}\label{sec:reference}

\bibliographystyle{lrec-coling2024-natbib}
\bibliography{lrec-coling2024-example}

\end{document}